\documentclass{article}
\usepackage[preprint]{neurips_2024}

\input{packages}
\newcommand{\model}{\texttt{ALLaM}}

\newcommand{\llamatwo}{\texttt{Llama-2}}

\title{\model{}: Large Language Models for Arabic and English}
\newcommand*\samethanks[1][\value{footnote}]{\footnotemark[#1]}
{
\author{M Saiful Bari \thanks{Corresponding author: skarim@sdaia.gov.sa} \ \thanks{Core contributors} \quad \textbf{Yazeed Alnumay}\samethanks\\ \textbf{Norah A. Alzahrani} \quad  \textbf{Nouf M. Alotaibi} \quad  \textbf{Hisham A. Alyahya}\\
 \quad \textbf{Sultan AlRashed} \quad \textbf{Faisal A. Mirza} \quad \textbf{Shaykhah Z. Alsubaie} \\
\textbf{Hassan A. Alahmed} \quad \textbf{Ghadah Alabduljabbar} \quad \textbf{Raghad Alkhathran} \\ \textbf{Yousef Almushayqih} \quad \textbf{Raneem Alnajim} \quad   \\ \quad \textbf{Salman Alsubaihi} \quad \textbf{Maryam Al Mansour} \quad\\  \textbf{Majed Alrubaian} \quad \textbf{Ali Alammari} \quad \textbf{Zaki Alawami} \\ \textbf{Abdulmohsen Al-Thubaity} \quad \textbf{Ahmed Abdelali}\quad \textbf{Jeril Kuriakose} \\ \textbf{Abdalghani Abujabal}\samethanks \quad \textbf{Nora Al-Twairesh}\samethanks \quad \textbf{Areeb Alowisheq}\samethanks\quad \textbf{Haidar Khan}\samethanks
\\\\
National Center for AI (NCAI), Saudi Data and AI Authority (SDAIA) \\ Riyadh, Saudi Arabia
}

\begin{document}
\maketitle

\begin{abstract}%
We present \model{}: \textbf{A}rabic \textbf{L}arge \textbf{L}\textbf{a}nguage \textbf{M}odel, a series of large language models to support the ecosystem of Arabic Language Technologies (ALT). \model{} is carefully trained considering the values of \emph{language alignment} and \emph{knowledge transfer} at \emph{scale}. Our autoregressive decoder-only architecture models demonstrate how second-language acquisition via vocabulary expansion and pretraining on a mixture of Arabic and English text can steer a model towards a new language (Arabic) without any catastrophic forgetting in the original language (English). Furthermore, we highlight the effectiveness of using parallel/translated data to aid the process of knowledge alignment between languages. Finally, we show that extensive alignment with human preferences can significantly enhance the performance of a language model compared to models of a larger scale with lower quality alignment. \model{} achieves state-of-the-art performance in various Arabic benchmarks, including MMLU Arabic, ACVA, and Arabic Exams. Our aligned models improve both in Arabic and English from their base aligned models.

\end{abstract}

\newpage
\setlength{\cftbeforesubsecskip}{10pt}
\setlength{\cftbeforesecskip}{10pt}
{\tableofcontents} %
\newpage

\section{Introduction}
\label{sec:intro}
Language modeling has significantly progressed from its humble origins, transitioning from fundamental probabilistic methods to complex neural priors. The foundational work by \citet{shannon1951prediction} on the information theory of language laid the groundwork for predicting the next word in a sequence, which was subsequently tackled by \citet{Bengio2003ANP} in neural networks. The field experienced a substantial leap with the introduction of LSTMs \citep{lstm} in language models (LM) \citep{peters2018deep}, which could capture longer dependencies in LMs but proved difficult to scale. The emergence of scalable and distributed architectures like Transformers \citep{vaswani2023attention} and the potential for precisely \citep{kaplan2020scaling,chincila} compressing web-scale data has resonated in recent years with the advancements of \emph{Generative Pretraining} \citep{gpt,brown2020language,anil2023palm}.

With the release of \texttt{ChatGPT} \citep{chatgptblog}, followed by the introduction of more frontier class models \texttt{Gemini} \citep{geminiteam2024gemini}, \texttt{Claude} \citep{claude3}, \texttt{Reka} \citep{rekateam2024reka}, \texttt{Mistral} \citep{mistralnews}, \texttt{Llama-3} \citep{llamablog} and recently released \texttt{Qwen-2} \citep{qwen2}, large language models have demonstrated significant leaps over each generation of models \citep{laskar2023systematic}. This exponential growth in performance has raised hope in the possibility of achieving Artificial General Intelligence \citep{hendrycks2022xrisk,marcusblog}. This rapid advancement has spurred discussions across various fields, including ethics, economics, and technology \citep{weidinger2021ethical}. Judging from the initial capabilities \citep{bubeck2023sparks}, the potential of these frontier models are reinventing the way humans interact with machines, impacting social norms, productivity, trends, and culture on a broader scale \citep{zhou2024real}. However, most of these frontier-class models are primarily trained on English and often lack a connection to localized regional cultures and norms \citep{naous2024having}. This gap has the potential to result in slow and irreversible manipulation of regional identities and lead to cultural homogenization.

The natural course to reverse this trend is to invest resources in curating data and building models to support the diversity of languages and cultures represented in the modern world. While this is possible, the significant training costs of LLMs and their environmental impact have become major concerns in recent years \citep{strubell2019energy}. The vast computational resources required to train LLMs contribute to substantial carbon emissions \citep{luccioni2023counting}. Governments \footnote{\url{https://www.cnrs.fr/en/update/jean-zay-supercomputer-recycling-its-heat}} and non/for-profit organizations \citep{dodge2022measuring,google_carbon_free,aws_carbon_free}, are increasingly aware of these issues. This awareness has led to discussions about the ethical implications of AI development and the need for sustainable practices concerning “\emph{When and how to scale the training of these models}”. In addition, curating data for each language/region at pretraining scale is also an extremely difficult task, since most available data comes from a few high-resource languages. 

\begin{figure*}
    \centering
    \includegraphics[width=\textwidth]{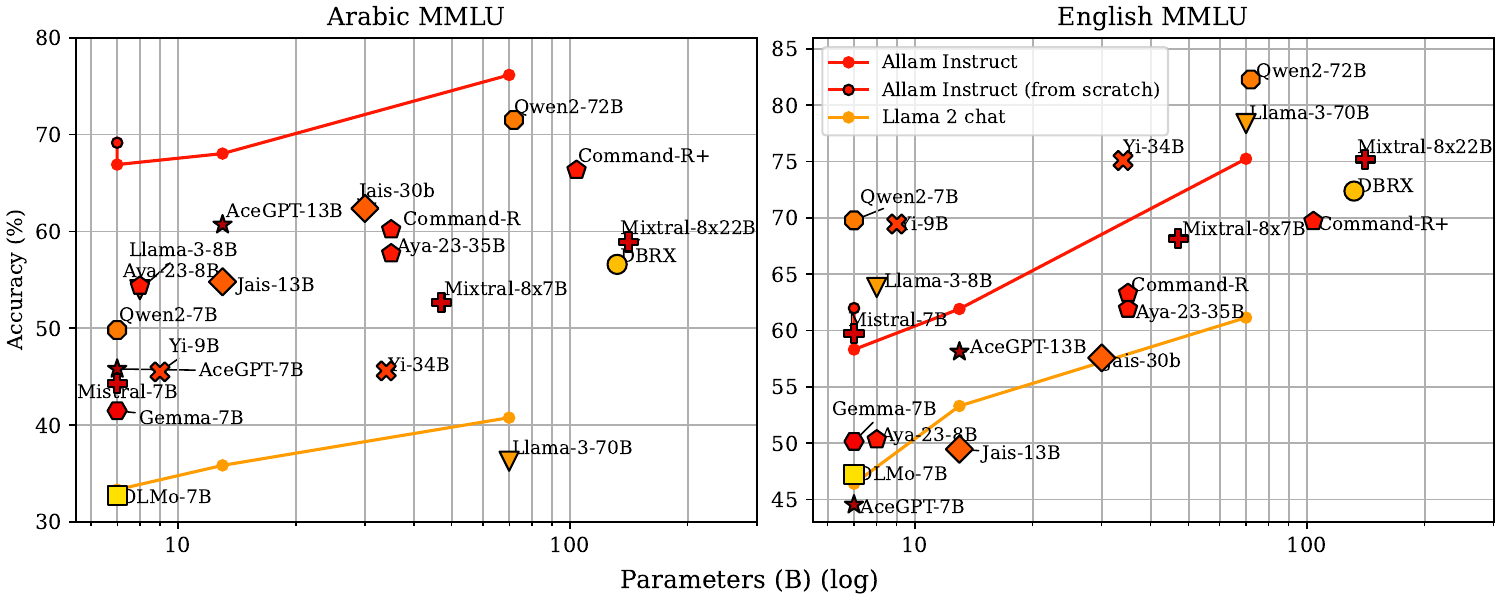}
    \caption{Performance on Arabic \citep{arabicmmlumbzu} and English \citep{mmlu} MMLU Benchmarks. \model{} (red line) shows impressive improvement from its base model \llamatwo{} (yellow line). All evaluations were done on the latest version of the fine-tuned (chat or instruct) models. The \model{} 7B from scratch model also shows significant improvement over the \model{} 7B continued pretraining model.}
    \label{fig:other_model_evals}
\end{figure*}

\begin{wrapfigure}{r}{0.5\textwidth}
    \vspace{-2em}
  \begin{center}
    \includegraphics[width=\linewidth]{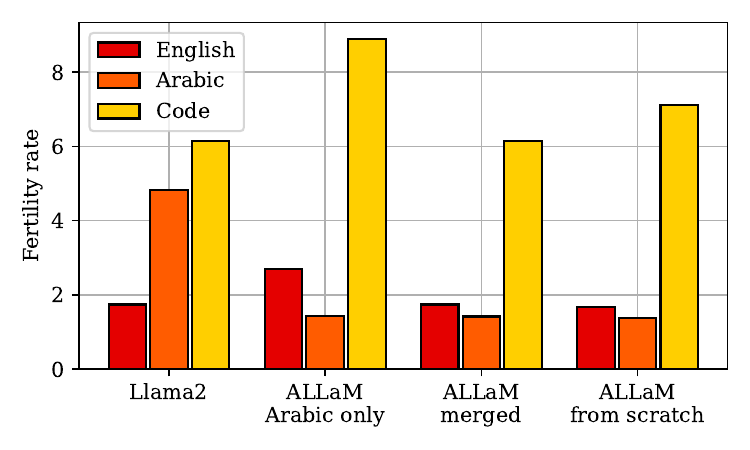}
  \end{center}
  \caption{Comparison of tokenizer fertility rates. The chart illustrates the fertility rates across four tokenizers: \llamatwo{}, \model{} Arabic only, \model{} merged with \llamatwo{}, and \model{} Arabic/English (from scratch model). We calculate the fertility over a random subsample of the entire English, Arabic, and code training corpus.}
    \label{fig:fertility}
  \vspace{-2em}
\end{wrapfigure}

To address these concerns, we consider the problem of adapting strong, but potentially under-trained, open pretrained models, rather than starting from a randomly initialized model. Technically, this involves continuing training of a model in a new language to facilitate Second Language Acquisition (SLA) \citep{MERRILL95}, popularized by \citet{bari2020zero} in NLP and recently adapted to LLMs by \citet{nguyen2023seallms}. This process involves the challenging task of incorporating an additional language distribution without compromising the source language(s). For instance, if a pretrained model was initially trained in English, expanding to an additional language presents challenges related to tokenization. \Cref{fig:fertility} gives an overview of \model{}'s tokenizers compared to a tokenizer primarily trained on English.

Our approach to building \model{}, large language models developed specifically for fluency and understanding in Arabic and English, can be outlined as follows. We first demonstrate the feasibility of adapting an existing pretrained English model (\llamatwo{} \citep{touvron2023llama2}) to fluency in both Arabic and English through \textit{tokenizer and vocabulary expansion}. Then, we apply our learnings to train a stronger model from scratch (random initialization) in a similar fashion, i.e., pretraining on English followed by training in mixed Arabic and English. The resulting model exhibits impressive performance and has favorable tokenization properties compared to other models.
This approach aligns with both our technical goals and our commitment to sustainable practices.

Our overall contributions are summarized below:
\begin{itemize}
\setlength{\itemsep}{0pt}
\setlength{\parsep}{0pt}
\item The \model{} model series, with the goal of supporting the cultural values of the Arabic-speaking world. We train four models at three different scales: 7B, 13B, and 70B models initialized by \llamatwo{} weights and a 7B model from scratch/random initialization.
\item Our model achieves state-of-the-art results in Arabic, as well as improving overall English performance of the original \llamatwo{} model. Refer to \Cref{fig:other_model_evals} for an overview.
\item We demonstrate that it is possible to train highly-performant models in low-resource languages from publicly available model weights using our continued pretraining recipe with tokenizer expansion, presenting a path for better representation of low-resource languages.
\item The training methodology and decision-making involved in training the LLM. We provide necessary ablation studies for most crucial decisions.
\end{itemize}

\section{Pretraining}
\label{sec:pretraining}
\begin{figure*}
    \centering
    \includegraphics[width=\textwidth]{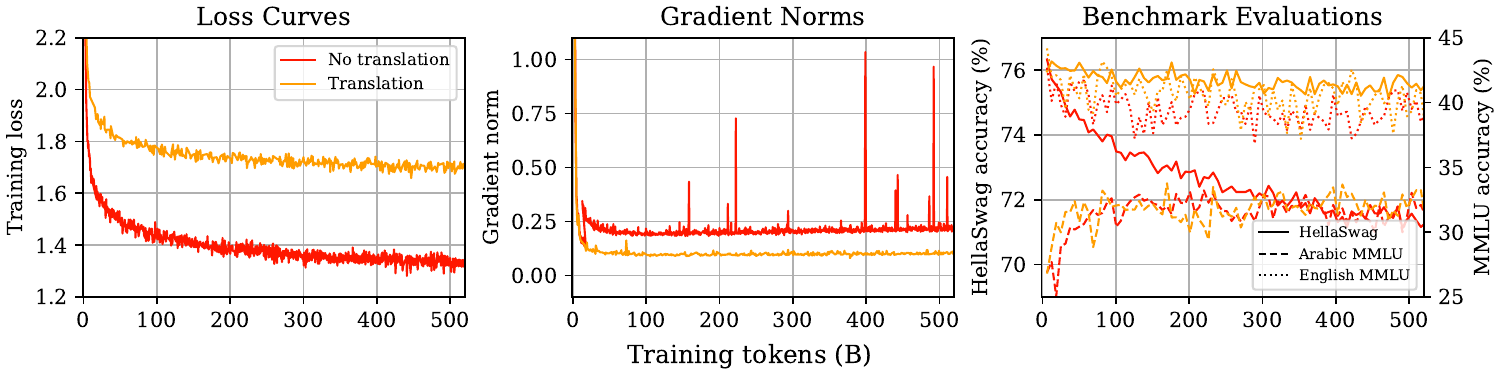}
    \caption{Measuring the effect of adding machine translated Arabic data to pretraining. Although the two loss curves look normal (\textit{left}), adding the translated Arabic reduced the frequency of gradient spikes during training (\textit{center}). Adding translated Arabic data also clearly helps align the Arabic and English capabilities of the model and reduce catastrophic forgetting (\textit{right}).}
    \label{fig:translation_effect}
\end{figure*}

Pretraining language models on trillions of natural language tokens represents the bulk of the cost required to build an effective language model. This large investment of time and compute precludes experimentation or ablation for every decision. Thus, before starting to train \model{} from random initialization, or “scratch”, we experiment in the continue-pretraining regime. As the name implies, continue pretraining is the practice of warm-starting a pretraining experiment from an already pretrained LM. 

We begin by discussing our entire pretraining corpus, describe experiments conducted with continue-pretraining, and finally describe pretraining from scratch.

\subsection{Pretraining Data}\label{sec2}

For English, many high quality and large scale datasets are available for pretraining \citep{together2023redpajama,DBLP:journals/corr/abs-2402-00159,DBLP:journals/corr/abs-2101-00027,refinedweb}. We harnessed subsets from Dolma-v1~\citep{DBLP:journals/corr/abs-2402-00159} and Pile~\citep{DBLP:journals/corr/abs-2101-00027} datasets e.g., Dolma-CC, The Stack \citep{kocetkov2022stack}, PeS2o, PubMed, DM-Math \citep{saxton2019analysing} and StackExchange \citep{cerebras2023slimpajama}. In total, we had access to 4T high to medium quality English tokens for pretraining.

Pretraining data in the Arabic language is much more limited, thus we undertook large scale collection and curation of Arabic language data.
This includes in-house crawled sources covering Web documents, news articles, books (literature, religion, law and culture, among others), Wikipedia (over 1M articles), and audio transcripts (books and news)\footnote{We are currently working on systematic auditing of our pretraining data. We do not have any timeline or visibility on when or if we can share our data for research.}. To ensure high quality Web data, we applied the following processing steps: \emph{(i)} drop documents with language identification score below 95\%,
\emph{(ii)} drop short documents that are less than 30 words, \emph{(iii)} drop documents with duplicate URLs or high ratio of spam and stop words,
and \emph{(iv)} drop duplicate documents using exact matching. We experimented with fuzzy matching but opted against using it as it was too restrictive given the scarcity of Arabic data.

\begin{wrapfigure}{r}{0.5\textwidth}
    \vspace{-2em}
  \begin{center}
    \includegraphics[width=\linewidth]{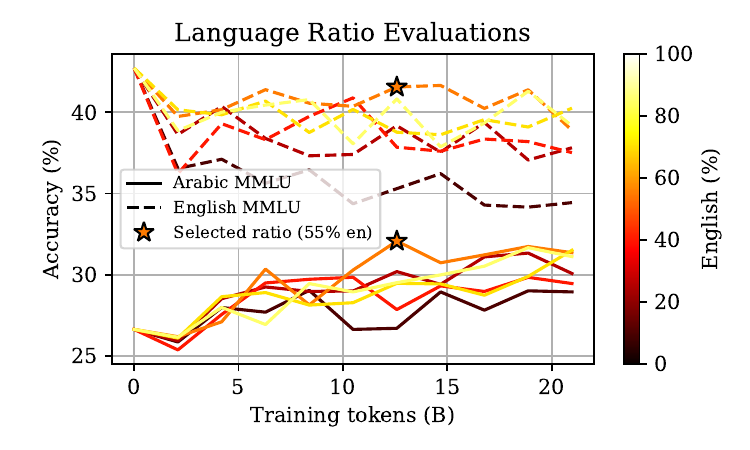}
  \end{center}
  \caption{We determine the optimal Arabic/English language mixture that balances between acquiring Arabic understanding while retaining English proficiency by conducting ablations over 6 Arabic/English ratios (trained up to 20B tokens). We found that a 45/55 Arabic/English ratio achieves the best performance, as measured by English and translated Arabic MMLU.}
    \label{fig:lang_ratio}
  \vspace{-2em}
\end{wrapfigure}

Additionally, we extended our Arabic data with translated English content using an in-house machine translation system. We translated the following English datasets from Dolma: Wikipedia, books, C4 and peS2o, which also are part of our English data. The hypothesis is that this will improve Arabic-English language alignment, leading to a better Arabic model. \Cref{fig:translation_effect} demonstrates the impact of Arabic translated datasets in the pretraining data mixture. While models trained without translated data exhibit lower training loss, those trained with translated data show more stable training, as evidenced by fewer spikes in gradient norms. Incorporating Arabic translated data in the pretraining dataset mitigates catastrophic forgetting in English. In total, we curate 540B Arabic tokens\footnote{Tokens counted by our merged tokenizer.} of which 270B are natural Arabic tokens and 270B are translated Arabic tokens.

\paragraph{Data Mixture}
To build a performant model in both Arabic and English, we conducted experiments to determine the optimal language mix. \Cref{fig:lang_ratio} gives an overview of data-mixture experiments on our curated Arabic-English corpus. We conducted the experiments with the same sampling ratio (\Cref{table:data_distribution}) and data order. We observe best trend in performance with 45/45 Arabic/English data mix.

\Cref{table:data_distribution} shows the language and category mixing distributions for English, Arabic natural, Arabic translated and final mix. Following mainstream work \citep{touvron2023llama,palm,DBLP:journals/corr/abs-2112-11446}, web data constitutes the highest ratio with 71\%, 65\% and 48\% of the Arabic natural, Arabic translated and overall mix, respectively. We limited the contribution of English web data to 31\%, as the \llamatwo{} base model was trained on a significant amount of English web data already, and we expected that increasing its ratio might degrade performance. We ensured that high quality sources, such as books, news articles, and code, are well represented in our mixture.

\begin{table}
\centering
\resizebox{.65\columnwidth}{!}{%
\begin{tabular}{lccccc}\toprule
\multirow{3}{*}{\textbf{Domain}} & \multicolumn{4}{c}{\textbf{Mixed Arabic \& English}} & \multirow{3}{*}{\textbf{English Only}} \\
\cmidrule(lr){2-5}
& \multirow{2}{*}{\textbf{English}} & \multicolumn{2}{c}{\textbf{Arabic}} & \multirow{2}{*}{\textbf{Mixed}} & \\\cmidrule(lr){3-4} 
& & \textbf{Natural} & \textbf{Translated} & \\\midrule
Web     & 31\% & 71\%   & 65\%   & 48\% & 71\% \\ 
Books   & 9\%  & 13\%   & 12\%   & 11\% & 3\% \\ 
Wiki    & ---  & 0.70\% & 0.61\% & 0.3\% & 0.1\% \\ 
News    & ---  & 14\%   & ---    & 3\% & --- \\ 
Science & 16\% & ---    & 22\%   & 14\% & 6\%\\ 
Code    & 39\% & ---    & ---    & 21\% & 17\% \\ 
Math    & 5\%  & ---    & ---    & 2.5\% & 0.9\% \\ 
Other   & ---  & 1.3\%   & 0.39\%    & 0.2\% & 2\% \\ \midrule
Lang Mix   & 55\% & 22.5\% & 22.5\% & 100\% & 100\% \\ 
\midrule
Tokens   & 660B & 270B & 270B & 1.2T & 4T \\ 
\bottomrule
\end{tabular}%
}
\vspace{1em}
\caption{\model{}'s pretraining data mixtures. The first four columns summarize the distribution of the continued pretraining mixed Arabic/English data. The English only pretraining from scratch mixture is shown in the last column. We upsample data to match the mixture rates when needed.}
\label{table:data_distribution}
\end{table}

\subsection{Continued Pretraining}

Open source and open weight models present an attractive option to conduct pretraining experiments cheaply. However, they also present challenges, since most such models do not natively support Arabic or other languages. We develop a simple approach to enhance any language model with capabilities in new languages (i.e., language expansion). The approach relies on two steps: \emph{(i)} tokenizer augmentation, and \emph{(ii)} expanded vocabulary learning. We demonstrate that this approach leads to minimal degradation of capabilities in the original language.

\paragraph{Tokenizer Augmentation}
Existing open weight language models (e.g., \llamatwo{}) tokenize Arabic (and other languages) poorly, often splitting words down to the character level or even relying on byte-fallback mechanisms for tokenization. This results in: \emph{(i)} inefficient training, as the pretraining corpus size is inflated, \emph{(ii)} unoptimized inference, since the model must generate more tokens per word, and \emph{(iii)} the effective context length is reduced, because it is based on a fixed number of tokens. To address these issues, we use a corpus of text in the target language to train a tokenizer specialized in that language. We then merge the original tokenizer with the language-specific tokenizer. Merging is accomplished by adding all tokens from the language-specific tokenizer that do not exist in the original tokenizer. As shown in \Cref{fig:fertility}, this effectively reduces the fertility rate in the target language of the merged tokenizer to the level of the language-specific tokenizer. 

\begin{figure}[th]
    \centering
    \begin{minipage}[t]{0.475\textwidth}
        \centering
        \vspace{0pt} % Align at the top
        \includegraphics[width=\linewidth]{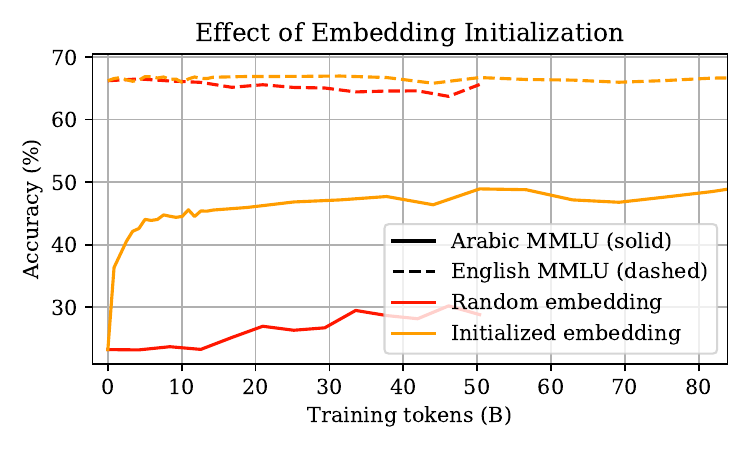}
        \caption{Effect of \emph{random initialization} vs. \emph{embedding initialization} during the start of continued pretraining. We find that initializing the embeddings for new tokens from combinations of existing embeddings speeds up learning dramatically.}
        \label{fig:init_effect}
    \end{minipage}\hfill
    \begin{minipage}[t]{0.475\textwidth}
        \centering
        \vspace{0pt} % Align at the top
        \includegraphics[width=\linewidth]{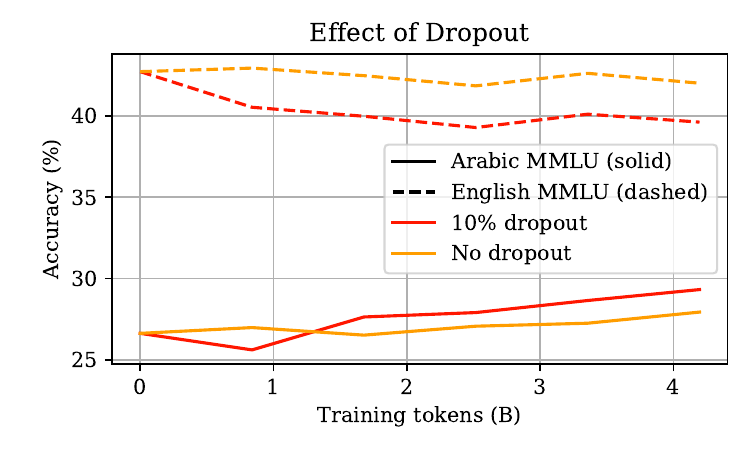}
        \caption{Effect of \emph{dropout} during the start of continued pretraining experiments. While introducing dropout can marginally improve the second language acquisition, it negatively impacts the model's capabilities in the original language.}
        \label{fig:dropout_effect}
    \end{minipage}
\end{figure}

\paragraph{Expanded Vocabulary Learning}
Newly added tokens in the merged tokenizer have no associated embedding representations in the pretrained language model's weights. To learn these representations, we experiment with two approaches: \emph{(i)} random initialization and \emph{(ii)} initialization from combined representations of tokens in the original tokenizer. Approach \emph{(ii)} is accomplished by tokenizing each token in the vocabulary of the new tokenizer using the original tokenizer. The associated embedding representations of this tokenization are then averaged and assigned as the vector representation of the new token. Since we work with tokenizers with byte-fallback, such a tokenization is guaranteed to exist. \Cref{fig:init_effect} provides an overview of our initialization method. Initializing the new embeddings from the combination of previously learned embeddings gives a significant boost to the learning of a new language. 

\paragraph{Experiment Details}
Starting from \llamatwo{} pretrained model weights, we continue pretraining the \model{}-7B and \model{}-13B models on 1.2T tokens, covering both English and Arabic languages. For the \model{}-70B model, we only train up to 600B tokens (using the same data mixture). In all of our continued pretraining experiments, we used the final learning rate of the pretrained language model (usually $3\times10^{-5}$). We experimented with approaches to gradually increase and decay the learning rate with limited success, as such models typically exhibited catastrophic forgetting, indicated by significant drops in performance in the original language. We also considered optimizer state warm up, as open-weight models typically do not include the optimizer states, but found this had little effect on performance. \Cref{fig:dropout_effect} provides an overview of adding dropout during continued pretraining. We observe that adding dropout helps the Arabic language, as it acts as a regularizer for the new distribution. However, \llamatwo{} was pretrained on 2T tokens without any dropout, and adding dropout negatively impacts the source language performance. Considering this trade-off, we decided not to add dropout in the continued pretraining stage. Unlike recent trends ~\citep{olmo17}, we did not add any alignment data in this stage of training. 

\subsection{Pretraining From Scratch}

Following \citep{chincila,touvron2023llama2}, training a high-quality model from scratch requires a substantial amount of tokens. Even when pretraining from random initialization, we find it beneficial to train with a high-resource language for trillions of tokens (English) and then continue training with a mixture of Arabic and English tokens.
On small scale experiments (with 1B parameter models) we find that beginning training with two languages can sometimes degrade the performance in English or result in slow learning of both language distributions. From this, we hypothesize that low-resource languages are diluted in the large volume of high-resource language data when pretraining from scratch, even with upsampling and careful tuning.

\paragraph{Training Recipe}
Our pretraining from scratch recipe consists of two steps: training on 4T English tokens followed by training on 1.2T mixed Arabic/English tokens. This retains the English capabilities of the model without catastrophic forgetting, effectively transferring knowledge from one language distribution to another. The only difference between pretraining from scratch and continued pretraining from an existing model is that vocabulary expansion is not required.

We match hyperparameters and architecture for pretraining from scratch with \citet{touvron2023llama2}, including 4M tokens per batch and max LR $3\times10^{-4}$ decayed to $3\times10^{-5}$ with a cosine schedule.

\paragraph{English Data Mixture} 
The last column of \Cref{table:data_distribution} shows the domain mix for the English only pretraining dataset. As expected, web data represent the bulk of the mixture, followed by code and scientific articles.

\subsection{Compute and Training Infrastructure}

Over the course of our development of \model{}, we had access to 128-1024 A100 GPUs. Our GPU cluster was equipped with InfiniBand connections to enable high-speed communication between nodes. The all-reduce test on the cluster ranges around 1200-1400 Gbps (node-node interconnect (RoCE)). The entire training period of the models is estimated to be 5M GPU hours. 

At the start of the project, we forked Megatron-LM\footnote{\url{https://github.com/NVIDIA/Megatron-LM}} and applied our own customizations (including improving data iterators, adding metadata in the checkpoints, and custom data pipelines). We utilized data, tensor, and pipeline parallelism supported by Megatron-LM to efficiently train at a large scale as well as FlashAttention~\citep{dao2022flashattention,dao2023flashattention2}. By leveraging these techniques, we achieved significant improvements in training speed. The throughput per GPU varied from 135 to 167 TFlop/s/GPU depending on the number of GPUs, number of nodes, batch size, and parallelism strategy. We trained \model{} with \texttt{bf16} mixed-precision. 

\section{Alignment}
\label{sec:alignment}
Building useful LLMs requires ensuring they are able to follow instructions while adhering to ethical standards and user expectations. This alignment process is especially crucial for models used in diverse linguistic and cultural contexts. In our setting, this means aligning models to the Arabic language and cultural context while also supporting English.

Supervised Fine-Tuning (SFT) (\Cref{sec:sft}) refines a pretrained model using a carefully selected dataset relevant to specific tasks and domains. Preference training (\Cref{sec:dpo}), on the other hand, aligns the model's outputs with human values and preferences by prioritizing responses that meet user expectations and ethical guidelines. These methods work together to create reliable and ethically sound LLMs for real-world use.

\subsection{Supervised Fine-Tuning}
\label{sec:sft}

\paragraph{Data}
Our SFT data is curated from a diverse array of sources. Given a piece of context from a source, we utilize humans and/or generative models \citep{ding2023enhancing} to identify if the text can be considered suitable for supervised fine-tuning or if we can generate instructions to create an SFT example from the context. For English, we primarily use public web content as our main source, offering a broad range of high-quality and especially diverse prompts. In contrast, our Arabic data comes from a combination of public and proprietary sources to ensure comprehensive coverage and relevance. To gather data from the source, we collect seed websites or data sources, which involves utilizing domain experts, prompt librarians, local institutes specializing in areas such as Arabic language, history, and politics, the use of commercially permissible licensed LLMs to generate data, and machine translation models to convert rich English SFT data into Arabic. Our datasets cover various domains and capabilities, ensuring the model's proficiency in handling tasks across education, history, Arabic linguistics, politics, religion, computer science, and other fields. The entire Arabic/English collection is called \texttt{Ultra-Instinct}, which is not \emph{human generated}, but rather, \emph{human driven}.

\paragraph{Quality Filtering} 
Unlike \citet{zhou2023lima,ai2024yi}, we hypothesized that scaling SFT data can unlock diverse capability, as well as improve responsiveness to the prompts. Initially, we crawled the public web for SFT samples. The first version (v1) of \texttt{Ultra-Instinct} includes 12M samples evenly split between English and Arabic, while the second version (v2), is a reduced version with half the samples. Compared to v1, v2 underwent strict quality checks and human assessments of random subsamples. Our quality checks for v2 included \emph{(i)} assessments based on instruction/response word length, \emph{(ii)} lexical \footnote{Lexical diversity is calculated by taking the ratio of the total number of unique words to the total number of words across all samples, excluding stop words.} and semantic diversity, exact and near-exact lexical deduplication, \emph{(iii)} removal of low quality machine-translated Arabic data from English sources, and \emph{(iv)} ensuring diversity in questions and commands. 
For detailed metrics on instruction and response lengths and lexical diversity, see \Cref{table:sft_quality_metrics}. 

\begin{figure*}[t]
    \centering
    \begin{subfigure}[t]{0.32\textwidth}
        \centering
        \includegraphics[width=0.98\linewidth]{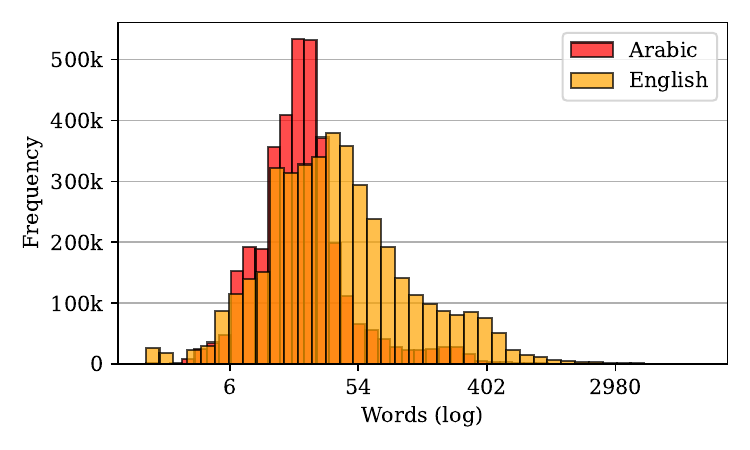}
        \caption{Prompt word count.}
        \label{fig:sft_prompt_word_distributions}
    \end{subfigure}
    \begin{subfigure}[t]{0.32\textwidth}
        \centering
        \includegraphics[width=0.98\linewidth]{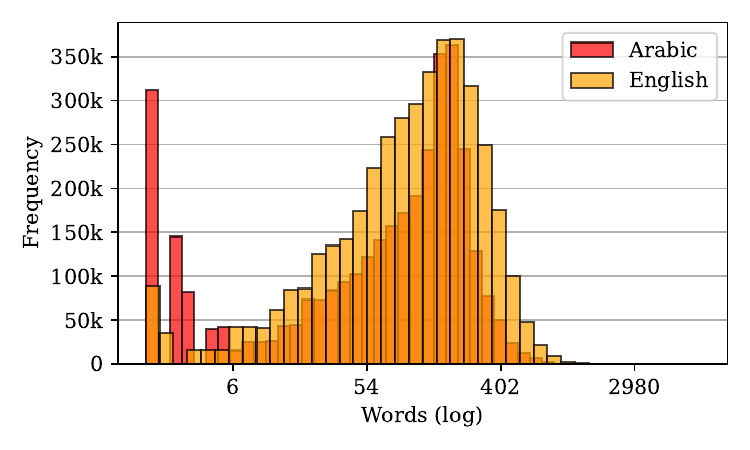}
        \caption{Response word count.}
        \label{fig:sft_response_word_distributions}
    \end{subfigure}
    \begin{subfigure}[t]{0.32\textwidth}
        \centering
        \includegraphics[width=0.98\linewidth]{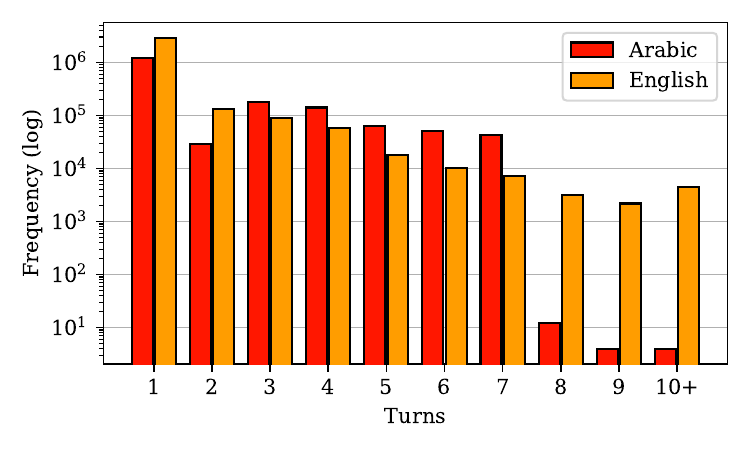}
        \caption{Number of conversation turns.}
        \label{fig:sft_chat_turns}
    \end{subfigure}
    \caption{Word count and turn distributions of SFT data. There are two main differences in our Arabic and English SFT datasets: shorter responses are more frequent in our Arabic SFT dataset, while our English SFT dataset contains more dialogues with >8 turns.}
\end{figure*}

\Cref{fig:sft_prompt_word_distributions,fig:sft_response_word_distributions} show the distribution of the prompts and responses in v2, respectively. We focused on maximizing the number of multi-turn conversations in our dataset. \Cref{fig:sft_chat_turns} shows the distribution of conversation turns from \texttt{Ultra-Instinct}.

\begin{table}
    \centering
    \resizebox{0.5\columnwidth}{!}{%
    \begin{tabular}{lcccc}\toprule
    \multirow{2}{*}{Quality Metric} & \multicolumn{2}{c}{\texttt{Ultra-Instinct} v1} & \multicolumn{2}{c}{\texttt{Ultra-Instinct} v2} \\
    \cmidrule(lr){2-3} \cmidrule(lr){4-5}
    & Prompt & Response & Prompt & Response \\\midrule
    Avg \# of Word     & 146.94 & 97.19   & 60.81  & 136.47 \\ 
    Lexical diversity   & 76.34  & 75.25   & 85.29  & 69.53 \\
    \bottomrule
    \end{tabular}%
    }
    \vspace{1em}
    \caption{Comparison of average word length and lexical diversity of prompts and responses.}
    \label{table:sft_quality_metrics}
\end{table}

\begin{table}[h]
\centering
\resizebox{0.75\columnwidth}{!}{%
\begin{tabular}{lcccccc}
\toprule
\multirow{2}{*}{Version} & \multicolumn{3}{c}{MMLU} & \multirow{2}{*}{Exams (ar)} & \multirow{2}{*}{ACVA} & \multirow{2}{*}{ETEC} \\
\cmidrule(lr){2-4}
 & \citet{acegpt} & \citet{arabicmmlumbzu} & en &  &  &  \\
\midrule
\texttt{Ultra-Instinct} v1 & 51.0 & 68.0 & 63.8 & 56.8 & 79.8 & 66.8 \\ 
\texttt{Ultra-Instinct} v2 & 51.4 & 68.5 & 63.3 & 56.8 & 76.7 & 65.9 \\ 
\bottomrule
\end{tabular}%
}
\vspace{1em}
\caption{Comparative results of \texttt{Ultra Instinct} v1 and v2, across various evaluation datasets.}
\label{table:sft_version_comparision}
\end{table}

To extrinsically evaluate the impact of higher quality SFT data, we trained two 13B models using our v1 and v2 SFT datasets. Even though v2 has half as many samples and v1, both versions performed equally well on English and Arabic evaluation benchmarks, as shown in \Cref{table:sft_version_comparision}. This reduction in data volume led to faster training times and reduced costs without compromising performance. It also clearly demonstrates the value of quality filtering for alignment.

\begin{wrapfigure}{r}{0.5\textwidth}
    \vspace{-2em}
    \centering
    \includegraphics[width=\linewidth]{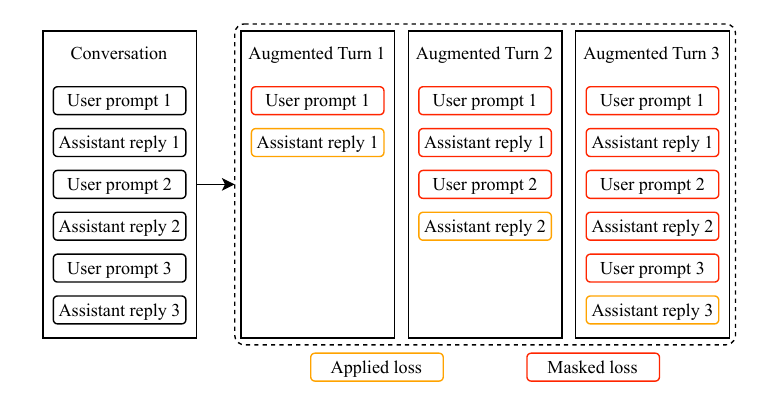}
    \caption{Augmentation process for conversations. The original conversation (left) is expanded into one sample per turn (right), with user prompts and assistant replies marked for training (red) and masking (orange) to enhance \model{}'s language understanding and multi-turn response generation capabilities.}
    \label{fig:turn_augmentation}
    \vspace{-1em}
\end{wrapfigure}

\paragraph{Training Details} 
We fine-tune our base model, which was trained on 3.2 trillion (2T \llamatwo{} + 1.2T \model{}) tokens, for 3 epochs using \texttt{Ultra-Instinct-v2} with a learning rate of $5\times10^{-6}$ and a batch size of 1024. The model is not trained to generate the prompt, as we mask out our prompt tokens when calculating the loss. \texttt{Ultra-Instinct-v2} contains a substantial number of multi-turn conversations. To train on these multi-turn conversations, we performed \texttt{turn-augmentation}. \Cref{fig:turn_augmentation} visually explains the process of turn augmentation. 

While training the SFT model, we encountered tokenization issues. Specifically, \llamatwo{}'s tokenizer was trained using \texttt{sentencepiece}\footnote{\url{https://github.com/google/sentencepiece}}, which breaks the beginning and end of sequence tokens into multiple tokens and adversely affects long multi-turn conversations. To address this issue, we patched \texttt{sentencepiece} using the \texttt{HuggingFace} \texttt{LlamaTokenizer} wrapper \citep{hf-transformers}. Over many iterations of training, we saw that even having 1\% noisy samples (e.g., empty responses or formatting issues) in alignment data can noticeably affect model quality.

\subsection{Preference Training}
\label{sec:dpo}

After SFT, models are able to converse in multi-turn conversations. However, they are not fully aligned with human preferences. For example, our SFT models were terse and had limited guardrails. To circumvent these issues, we performed preference tuning with human verified samples via Direct Preference Optimization (DPO) \citep{dpo}.

\paragraph{Data}
The inputs were sourced from early model testers and a manually curated selection of prompts from various domains or attack vectors. These include ethical dilemmas, middle eastern culture, religions, illegal activities, human rights, locale awareness, and personality.

Preference training necessitates both negative and positive outputs for each input. We relied on the testers' feedback to identify the positive outputs. In the absence of positive outputs, we used a model to generate an output and manually verified that the output was aligned. While \citep{zephyr} utilized preference data from AI Feedback (AIF) at scale, we adopt a more cautious approach in creating preference data. We generate a smaller volume of data, ensuring it is fully reviewed, edited, and verified by humans. 

There are two approaches for generating negative outputs: \emph{(i)} on-policy: use the generations of the model we are tuning as negative outputs, and \emph{(ii)} off-policy: use another similar model to generate the negative outputs. We did not verify that the negative outputs were worse than the positive. However, we ensured that the positive outputs were of the highest quality, such that they were almost always better than the negative outputs.

\citet{khan2023xcodeeval} demonstrated that model outputs can vary significantly depending on the sampling mechanism used. Building on this insight, we generate additional samples for each instance by varying temperature and nucleus sampling techniques. These additional samples are utilized to produce rejected samples, ensuring that \model{} provides more grounded responses and generalizes well across various sampling mechanisms. 

In total, we collected 25,854 samples (triplets of \{prompt, accepted, rejected\}) in English and Arabic language. Using the technique mentioned above, we sample 10 different response from the model to generate additional rejected responses for each sample. This results in a dataset of 245K samples (after filtering) for preference training.

\paragraph{Training}
For DPO, we used a batch size of 512 with $\texttt{KL}_{penalty}=0.1$ and a learning rate of $9\times10^{-7}$ decayed to $5\times10^{-7}$ using a cosine annealing learning rate schedule. We train \model{} for a single epoch using all the preference data.

From our initial experiments with small datasets, we observed issues with model quality even when a small fraction (0.1\%) of the data was noisy. In this context, noise can be improper labeling of positive/negative pairs or low quality positive outputs. It is not clear, however, if after scaling up the DPO data whether the model can ignore this type of noise. In early DPO models, trained on data where we did not verify all the samples, we found that even a few moderately noisy samples resulted in broken models that repeatedly generate the same text or output incoherent text.

\paragraph{DPO vs. PPO}
One of the fundamental differences between DPO and PPO is that PPO is always on-policy with an external reward model. In our experience with DPO, we did not encounter any significant issues with off-policy experiments. Additionally, DPO allows for faster iteration and easier understanding of the training dynamics. The decision to use DPO over PPO was based on logistical constraints rather than a performance comparison of the algorithms. Given our compute setup and time constraints, we chose to proceed with DPO. We plan to explore PPO in the future for alignment.

\section{Evaluation}
\label{sec:evaluation}
\begin{figure*}
    \centering
    \includegraphics[width=\textwidth]{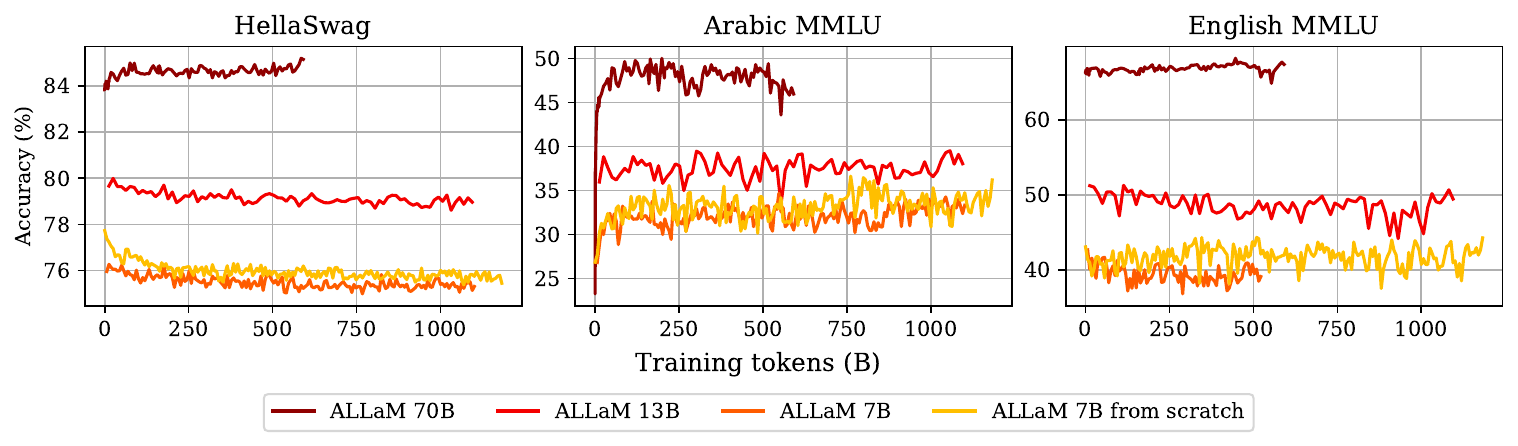}
    \caption{Selected benchmark evaluated through \model{}'s training. Using HellaSwag as a proxy for language understanding, the performance of smaller models degrades when introducing Arabic, while larger models (70B) have enough capacity to improve simultaneously in English and Arabic. Arabic language acquisition is rapid in all models, as indicated by Arabic MMLU.}
    \label{fig:evals_through_training}
\end{figure*}

\begin{table*}
    \centering
    \includegraphics[width=\textwidth]{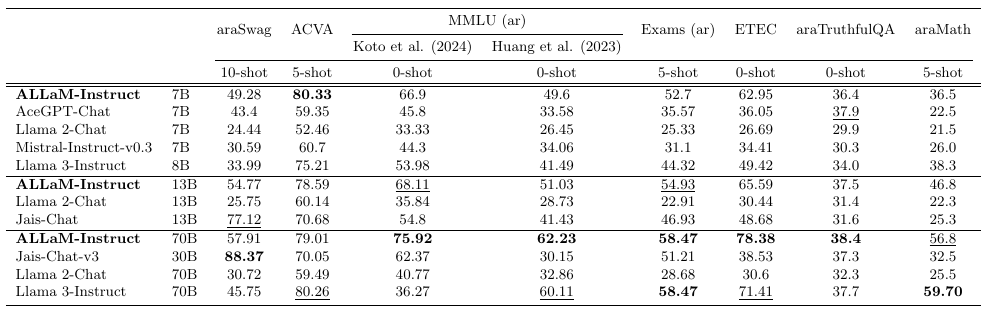}
    \caption{Arabic benchmark results for instruction tuned models.}
    \label{tab:instruct_only_ar}
\end{table*}

\begin{table*}
    \centering
    \includegraphics[width=\textwidth]{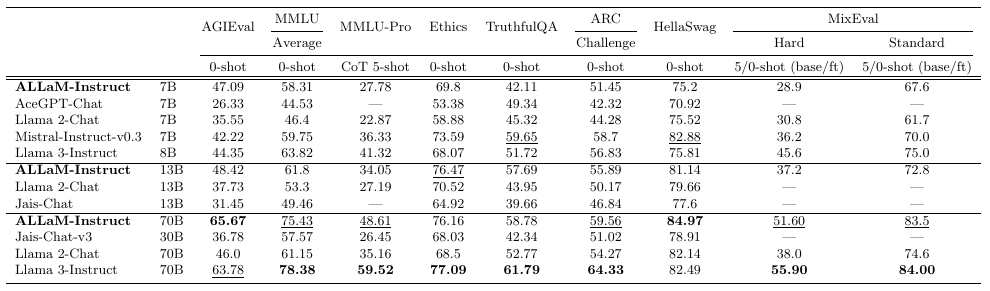}
    \caption{English benchmark results for instruction tuned models.}
    \label{tab:instruct_only_en}
\end{table*}

In this section, we describe the evaluation of our model and report the results of \model{} 7B, 13B, and 70B models, as well as other relevant models, such as \texttt{GPT-4}, \texttt{Command-R+} \citep{commandr}, and \texttt{Jais-30B} \citep{jais}. Our evaluations encompass three main types: \emph{(i)} automatic evaluations, \emph{(ii)} LLM-based evaluations, and \emph{(iii)} human evaluations.

\paragraph{Limitations}
Recently, \citep{alzahrani2024benchmarks} showed that multiple choice or cloze test based evaluation may not be robust. In addition, MT-Bench uses an LLM as a judge, and has likely leaked into training datasets. Unfortunately, human evaluation is time-consuming and requires well-trained human evaluators. In this work, we try to ensure robust evaluation and attain a balanced assessment of the quantitative metrics and qualitative effectiveness of models in various applications and domains.

\subsection{Automatic Evaluations}

The automatic evaluations cover Arabic and English benchmarks grouped into the categories listed below:
\begin{enumerate}
    \item Multi-domain: MixEval \citep{mixeval}, MMLU-Pro \citep{mmlupro}, and BBH \citep{bbh}.
    \item Reasoning and Commonsense: HellaSwag \citep{hellaswag},  PIQA \citep{piqa}, WinoGrande \citep{winogrande}, and AraSwag \citep{araswag}.
    \item World Knowledge and Language Understanding: MMLU \citep{mmlu},ARC Easy and Challenge \citep{arc},  TriviaQA \citep{triviaqa}, BoolQ \citep{boolq}, NQ Open \citep{nqopen}, AGIEval \citep{agieval}, Exams-Ar \citep{examsar}, MMLU Arabic (tr) \citep{acegpt}, MMLU Arabic (MBZU) \citep{arabicmmlumbzu} , and ETEC (in-house curated).
    \item Safety and Alignment: Hendrycks Ethics \citep{hendrycksethics}, ACVA \citep{acegpt}, TruthfulQA \citep{truthfulqa}, and AraTruthfulQA (in-house curated).
    \item Conversation: MT-Bench \citep{mtbench}, and Arabic domain capability dataset (in-house curated).
    \item Math: Minerva MATH \citep{minervamath, hendrycksmath}, GSM8K \citep{gsm8k} and AraMath (in-house curated).
    \item Coding: HumanEval \citep{humaneval-x}
\end{enumerate}

The following benchmarks were curated and developed in-house: 
\begin{itemize}
    \item ETEC: a collection of 1891 multiple choice questions covering different exams performed by the Education and Training Evaluation Commission in Saudi Arabia\footnote{\url{https://etec.gov.sa/home}}.
    \item AraMath: a set of 600 test samples that were post-processed and prepared from the AraMath~\citep{alghamdi-etal-2022-armath} dataset. These samples focus on testing the models' performance on Arabic math problems.
    \item AraTruthfulQA: a dataset created using similar methodology to the TruthfulQA~\citep{lin2021truthfulqa} dataset. It comprises a total of 541 samples, 285 samples were translated  directly from TruthfulQA using \texttt{GPT-4} and carefully validated and localized by human verifiers. Additionally, 256 questions were curated by humans to ensure their contextual relevance and cultural appropriateness.
\end{itemize}

While serving as a good test bench, observing the dynamics of automatic evaluations during training is also interesting. \Cref{fig:evals_through_training} shows the behavior of selected benchmarks during mixed Arabic/English pretraining while scaling up model size. In particular, we observe that smaller models tradeoff between capability in the new and original languages. However, larger models can simultaneously improve in both languages. 

Another observation from automatic evaluations is that some evaluations provide more signal for training decisions than others, e.g., Hellaswag smoothly improves during training while improvements in GSM8k occur in discontinuous jumps. Other benchmarks show no improvement until 1.5T tokens have been seen (i.e., grokking) making them unreliable for early training decisions.

\Cref{tab:instruct_only_ar,tab:instruct_only_en} give an overview of the performance of \model{} instruct models compared to other relevant models.

In Arabic benchmarks, we can see that \model{}-70B scores are the best in five (MMLU Arabic (natural and translated), Exams, ETEC, araTruthfulQA) out of the eight benchmark sets. On English, \model{} is the second-best model in the majority of cases, following \texttt{Llama 3-Instruct}. We highlight the excellent performance of \model{} on benchmarks released after training was completed (MMLU-Pro, MixEval) and benchmarks the training team did not have access to (ETEC), since they provide a clean evaluation signal.

\Cref{tab:ar_mega_table_appendix} and \Cref{tab:en_mega_table_appendix} in the appendix contain more detailed evaluation results for Arabic and English, respectively.

\paragraph{Evaluation Framework}
All evaluations were completed using the Language Model Evaluation Harness \citep{eval-harness} with the following exceptions: HumanEval was evaluated using BigCode Evaluation Harness \citep{bigcode-evaluation-harness}. MMLU-Pro, MixEval, and Arabic MMLU \citep{arabicmmlumbzu} were evaluated using the repositories of the dataset creators.

\subsection{LLM-Based Evaluations}

MT-Bench \citep{mtbench} consists of 80 multi-turn questions to evaluate models’ capabilities on complex instruction-following. In addition to the English version, MT-Bench Arabic was created using \texttt{GPT-4} to translate the original dataset and human annotators to review and align the prompts to Arabic culture. \texttt{GPT-4} serves as the LLM judge, scoring responses as recommended in \citep{mtbench}. Model performance is compared turn by turn, with results shown in \Cref{table:mt_bench}, where \model{}-70B achieves the best Arabic performance.

\begin{table}
\centering
\resizebox{0.75\columnwidth}{!}{%
\begin{tabular}{lcccccc}\toprule
\multirow{2}{*}{Model} & \multicolumn{3}{c}{English} & \multicolumn{3}{c}{Arabic}
\\\cmidrule(lr){2-4} \cmidrule(lr){5-7} 
& Avg. & Turn 1 & Turn 2 & Avg. & Turn 1 & Turn 2 \\\midrule

AceGPT 13B-chat  & 5.44 & 6.76 & 4.12 & 6.33 & 7.01 & 5.64  \\
ALLaM 13B Instruct & 7.34  & 7.67  & 7.01 & 7.57 & 7.9 & 7.23 \\
ALLaM 70B Instruct & 7.44 & \textbf{7.91} & 6.96 & \textbf{8.19} & \textbf{8.4} & \textbf{7.97}  \\
Jais 13B Chat & 4.18 & 4.39 & 3.96 & 4.72 & 5.07 & 4.36 \\
Jais 30B Chat v1 & 3.89  & 4.13 & 3.64 & 3.54 & 4.13 & 2.95 \\
Jais 30B Chat v3 & 5.86 & 6.25 & 5.47 & 6.28 & 6.78 & 5.78 \\
Cohere Command R+ & 7.41  & 7.63 & 7.18 & 7.97 & 8.28 & 7.65 \\
Cohere Command R & 6.99 & 7.19 & 6.79 & 7.47 & 7.82 & 7.12 \\
DBRX Instruct & 7.16 & 7.33 & 6.98 & 7.83 & 8.19 & 7.46 \\
GPT 3.5 Turbo & \textbf{7.55}  & 7.79 & \textbf{7.31} & 8.12 & 8.39 & 7.84 \\ \bottomrule
\end{tabular}%
}
\vspace{1em}
\caption{MT-Bench scores for Arabic and English. Each score is an average over 80 samples of the score ranging from 0 to 10 returned by the judge (\texttt{GPT-4}).}
\label{table:mt_bench}
\end{table}

\subsection{Human Evaluation}

We developed an Arabic multi-turn dataset that covers seven domains: Arabic linguistics, history, health, politics, coding, entertainment, and ethics, each domain contains ten questions with two turns. 

Human evaluators compared the responses from two models and were asked to choose the winning response with the following instructions:
\begin{itemize}
    \item Choose a response as the winner if it is the best, tie if both responses are equally good, and both-bad if both responses are not good.
    \item A response is considered good if it is coherent, grammatically correct, and is a reasonable response to the question or previous turn in the conversation.
    \item Good responses should be in the correct language (the response should be in the same language as the previous turn, unless another language was requested).
    \item Good responses should not contain toxicity, hate speech, or bias.
\end{itemize}

\begin{figure}[th]
    \centering
    \begin{minipage}[t]{0.45\textwidth}
        \centering
        \vspace{0pt} % Align at the top
        \includegraphics[width=\linewidth]{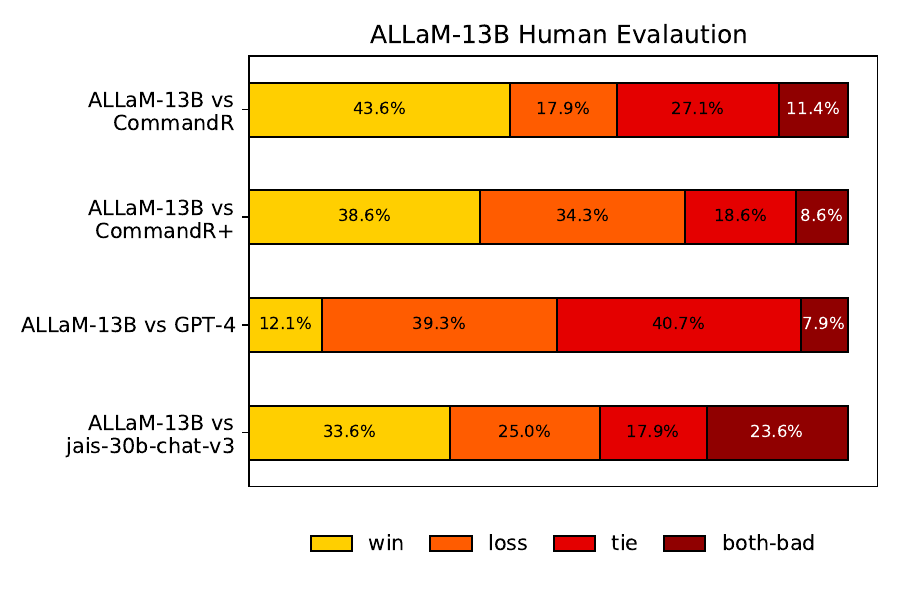}
        \caption{Pairwise win rates as judged by human evaluators. \model{}-13B wins against many much larger models.}
        \label{fig:allam-13-human-eval-models-compair}
    \end{minipage}\hfill
    \begin{minipage}[t]{0.45\textwidth}
        \centering
        \vspace{0pt} % Align at the top
        \includegraphics[width=\linewidth]{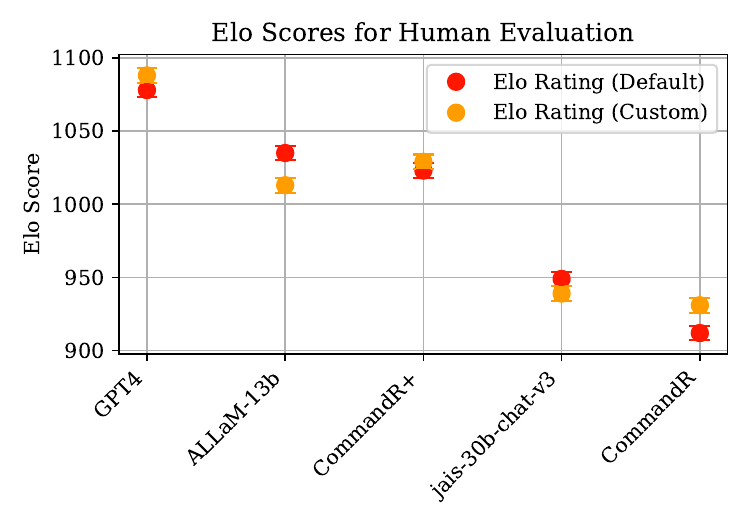}
        \caption{ELO scores from human evaluator preferences. \model{} is tied with \texttt{Command-R+} and lags only behind \texttt{GPT-4}.}
        \label{fig:elo-scores}
        \end{minipage}
\end{figure}

Each pair of responses was inspected by three evaluators, and the winner was determined by majority voting. In case of a tie, a fourth evaluator was used to break the tie. \Cref{fig:allam-13-human-eval-models-compair} presents the human evaluation results of the pair-wise comparisons of these models: \texttt{ALLaM-13b}, \texttt{Jais-30b-chat-v3}, \texttt{Command-R-plus}, and \texttt{Command-R-v01}. \texttt{ALLaM-13b}'s win rate was always higher than its loss rate compared with other models.

Finally, we gather votes from the human evaluators and calculated ELO scores for each model. ELO scoring had two configurations, the default scoring rewards the model for good responses with 1 point, tied responses (good and both-bad) with 0.5 points, and penalizes for bad responses with 0 point. The custom configuration penalizes the model with the bad response and both models if both provided bad responses with 0 point. \Cref{fig:elo-scores} shows the ELO scores based on the human evaluations. From the figure, \texttt{GPT-4} achieved the highest score, followed by \texttt{ALLaM-13b} with the second-highest score, outperforming (or matching) larger models such as \texttt{CommandR+}.

\section{Related Work}
\label{sec:related}
Our work sits at the cross-section of research building language models that support multiple languages and scaling such techniques in terms of size and data. To successfully train a large language model in a language other than English requires a complete understanding of cross-lingual transferability between languages and a good understanding of scaling laws as well as the fundamentals of training large language models. In this section, we discuss work on language modeling from the perspectives of cross-lingual alignment, multitask learning, and Arabic specialization.

\subsection{Language Modeling and Cross-lingual Representations}

In early work, word representations were derived using basic forms of the \textit{skip-gram model} \citep{mikolov2013distributed}, wherein each word is assigned a representation that does not account for varying contexts \citep{grave2018learning,GLOVE}. Further work in this area developed word representations that are adaptive to the context surrounding the words \citep{mccann2017learned,ELMO,howard-ruder-2018-universal,BERT,XLNet,radford2019language}.

\citet{ELMO} introduced \texttt{ELMo}, a model built with a bidirectional LSTM-based language model (LM) for pretraining to obtain contextualized word representations. This technique combines the outputs from all layers linearly when targeting specific tasks. Nonetheless, the sequential nature of LSTM-based LM pretraining presents challenges in scaling training efficiently. Concurrently, \citet{vaswani2023attention} developed the \textit{Transformer} architecture, which leverages multi-headed self-attention and positional encoding to handle long-range dependencies and enable parallel processing. Following that \citet{radford2019language} introduced \texttt{GPT}, a model that pretrains a Transformer decoder using a conditional language model objective, with subsequent fine-tuning requiring only minimal modifications. Similarly, \citet{BERT} unveiled \texttt{BERT}, which uses a Transformer encoder pretrained via a masked language modeling (MLM) objective. This approach excelled at task adaptation and benefited from the MLM’s ability to encode context bidirectionally, unlike the traditional unidirectional (conditional) LM that processes either the left or right context. Later \citet{t5} proposed a detailed hybrid encoder-decoder architecture based LLM with an implementation of many objective function via autoregressive structure. 

During the release of \texttt{BERT}, \texttt{mBERT}, a multilingual version of \texttt{BERT} is trained on 102 languages using a shared vocabulary of 110K subword tokens.\footnote{\url{https://github.com/google-research/bert/blob/master/multilingual.md}}. Despite the lack of explicit cross-lingual supervision, \texttt{mBERT} has demonstrated the ability to learn cross-lingual representations that generalize well across languages. \citet{xBERT,pires-etal-2019-multilingual} evaluated the zero-shot cross-lingual transferability of \texttt{mBERT} on several NLP tasks and attributed its generalization capability to shared subword units. \citet{pires-etal-2019-multilingual} additionally identified structural similarity (e.g., word order) as another crucial factor for successful cross-lingual transfer. \citet{k2020crosslingual}, however, argued that shared subwords contribute minimally, and instead, structural similarity between languages is more critical for effective transfer. \citet{artetxe2019crosslingual} further showed that joint training might not be necessary and proposed an alternative method to transfer a monolingual model to a bilingual model by learning only the word embeddings in the target language. They also highlighted the vocabulary size per language as an important factor. Finally, \citet{xue-etal-2021-mt5} showed that joint training on a large multilingual vocabulary can robustly map multilingual language models to the same latent space.

In the early days, \emph{Cross-lingual alignment} from mono-lingual embeddings was tricky and often required complex adversarial training \citep{MUSE}, careful orthogonal mapping \citep{artetxe2018aaai} or semi-supervised learning \citep{LNMap,bari2020zero}. With the introduction of \texttt{mBERT}, it became evident that learning joint distribution makes it easier for LLM to achieve cross-lingual alignment at scale. \citet{XLM} enhanced \texttt{mBERT} by incorporating a conditional LM and a translation LM (leveraging parallel data) objective along with a language embedding layer and trained a larger model utilizing more monolingual data. \citet{huang-etal-2019-unicoder} suggested employing auxiliary tasks like cross-lingual word recovery and paraphrase detection for pretraining. Subsequent work by \citet{XLMR} and \citet{soltan2022alexatm20bfewshotlearning} scaled up the training of multilingual language models. As well, \citet{xue-etal-2021-mt5} scale the size and languages in the T5 architecture. In an effort to reproduce GPT-3, \citet{BLOOM} trained the first auto-regressive multilingual LLM.

\subsection{Multitask Learning And Alignment}

Early work has demonstrated that multitask learning can enhance the performance of NLP models \citep{colobert}. In explicit multitask learning, augmenting all samples during training may introduce noise due to differing output distributions in a traditional full-model fine-tuning setup \citep{weller-etal-2022-use,bari2021}. For implicit multitask learning, \citet{radford2019language} showed that a language model can begin to learn downstream tasks without explicit supervision by pretraining alone. Large language models \citep{brown2020languagemodelsfewshotlearners,turing-nlg,palm} at scale can perform few-shot in-context learning, making them effective multitask models. Additionally, \citet{T0,FLAN,bloomz,flan_t5} found that these implicitly learned language models could be further improved by explicitly fine-tuning them with human instructions and prompts \citep{promptsource,super_natural} in a multitask fashion. Unlike previous template-based prompting approaches, \citet{ouyang2022traininglanguagemodelsfollow} applied preference tuning with reinforcement learning \citep{rlhf_summarization} using naturally written prompts. Subsequently, \citet{bai2022constitutionalaiharmlessnessai} introduced \emph{Constitutional AI} to automate alignment using AI feedback. Recently, following the work of \citet{rafailov2023directpreferenceoptimizationlanguage}, various efforts \citep{azar2023generaltheoreticalparadigmunderstand,ethayarajh2024ktomodelalignmentprospect,hong2024orpomonolithicpreferenceoptimization,park2024disentanglinglengthqualitydirect,meng2024simposimplepreferenceoptimization} have been directed towards preference tuning without explicit reward models.

% Language models for Arabic
\subsection{Language Models for Arabic}
As of the time of writing, the most prominent Arabic-focused LLMs are:
\begin{enumerate}
    \item \texttt{Jais} \citep{jais}: 13B and 30B base and chat models trained from scratch using a combination of natural and translated Arabic data along with English and code data.
    \item \texttt{AceGPT} \citep{acegpt}: 7B and 13B base and chat models trained from \llamatwo{} \emph{without} vocabulary expansion. 
    % They also highlight the dangers of using translated data on LLM localization.
\end{enumerate}

While \texttt{Jais} and \texttt{AceGPT} are currently the most prominent models, early open models such as \texttt{AraGPT} \citep{antoun2020aragpt2}, AraT5 \citep{elmadany2022arat5}, \texttt{AraBART} \citep{eddine2022arabart}, and \texttt{Noon} \citep{lakim2022holistic} \footnote{\url{https://huggingface.co/Naseej/noon-7b}} pioneered the area with models developed with limited resources to serve Arabic.

Other models such as \texttt{Jasmine} \citep{abdul2023jasmine} and \texttt{Aramus} \citep{alghamdi2023aramus} also showed the need for building a language model for over 400 million speakers worldwide. 

In addition to the language adaptation of models and multilingual models reviewed above, recent work has focused on building multilingual/bilingual language models from open weight language models. For example, \citet{rucinski2024efficient} adapted Mistral 7B for the Polish without vocabulary expansion. Mala-500 is another effort to expand to 534 languages by expanding the vocabulary to 260K tokens and further pretrained \llamatwo{} using LoRA adaptors \citep{lin2024mala}. Due to the number of languages they aimed to support, a small amount of data was included for each language and the evaluation of the approach was limited to measuring perplexity and automatic classification benchmarks. \citep{cui2023efficient} introduced a Chinese Language adaptation of \texttt{Llama} and \texttt{Alpaca} models, where the vocabulary was increased to 50K tokens, then continued to pretrain the models and finally fine-tune them.

\section{Conclusion}
\label{sec:conclusion}
The \model{} model series marks a significant advancement in Arabic Language Technologies, achieving state-of-the-art performance across various Arabic benchmarks while maintaining or enhancing English performance. Through careful training that emphasizes language alignment and transferability, our models demonstrate effective second-language acquisition without catastrophic forgetting. The strategic use of translated data, knowledge encoding, and alignment with human preferences have been crucial in this success.

\section{Limitations}

\model{} was trained on data that may potentially include toxic language, unsafe content, and societal biases originally sourced from the internet, leading to the possible amplification of these biases and toxic responses. Although \model{} underwent comprehensive safety training during the alignment phase, more community feedback is needed to iteratively improve \model{}. Additionally, inherent uncertainties in generative models mean that trials cannot encompass every possible use case, making it impossible to predict the model's responses in all contexts. This can occasionally result in inaccurate, biased, or socially unacceptable outputs, even if the prompt itself is not explicitly offensive. Developers must conduct thorough safety evaluations and make specific adjustments to ensure that \model{} is suitable for their intended purposes. Furthermore, the output generated by \model{} should not be considered a statement from \model{}'s creators or any affiliated organization.

\section{Ethical Statement}

While conducting and presenting this research, we are committed to upholding the highest ethical standards. We recognize the potential impact of large language models on society and the importance of ensuring their responsible development and deployment. Our work adheres to principles of fairness, transparency, and inclusivity, striving to mitigate biases and ensure diverse representation in our training data. We are mindful of privacy concerns and have taken steps to anonymize and secure data used in our research. Additionally, we acknowledge the potential for misuse of language technologies and advocate for their ethical application, promoting beneficial use cases while being vigilant about unintended consequences. \model{} models are made openly available to foster collaboration and further research, with the aim of contributing positively to the advancement of language technologies and supporting the cultural and technological growth of the Arabic-speaking world.

\section{Risk Statement}

The deployment and use of LLMs in various applications poses significant risks, including data privacy and security concerns due to the inadvertent inclusion of sensitive information in training datasets. LLMs can perpetuate or amplify biases, resulting in unfair treatment and discrimination in critical decision-making processes. They can also generate convincing but inaccurate content, spreading misinformation and potentially influencing public opinion negatively. Over-reliance on LLMs may diminish human judgment, and the models' susceptibility to adversarial attacks can compromise system integrity. To mitigate these risks, we follow robust governance, continuous monitoring, and iterative improvements. We also adhere to best practices in data handling and model training, fostering transparency and accountability in LLM development.

\bibliographystyle{custombibstyle}
\bibliography{custom}

\newpage
\appendix
\section{Appendix}
\subsection{Detailed Arabic Evaluation}
\begin{table*}[h]
    \centering
    \includegraphics[width=\textwidth]{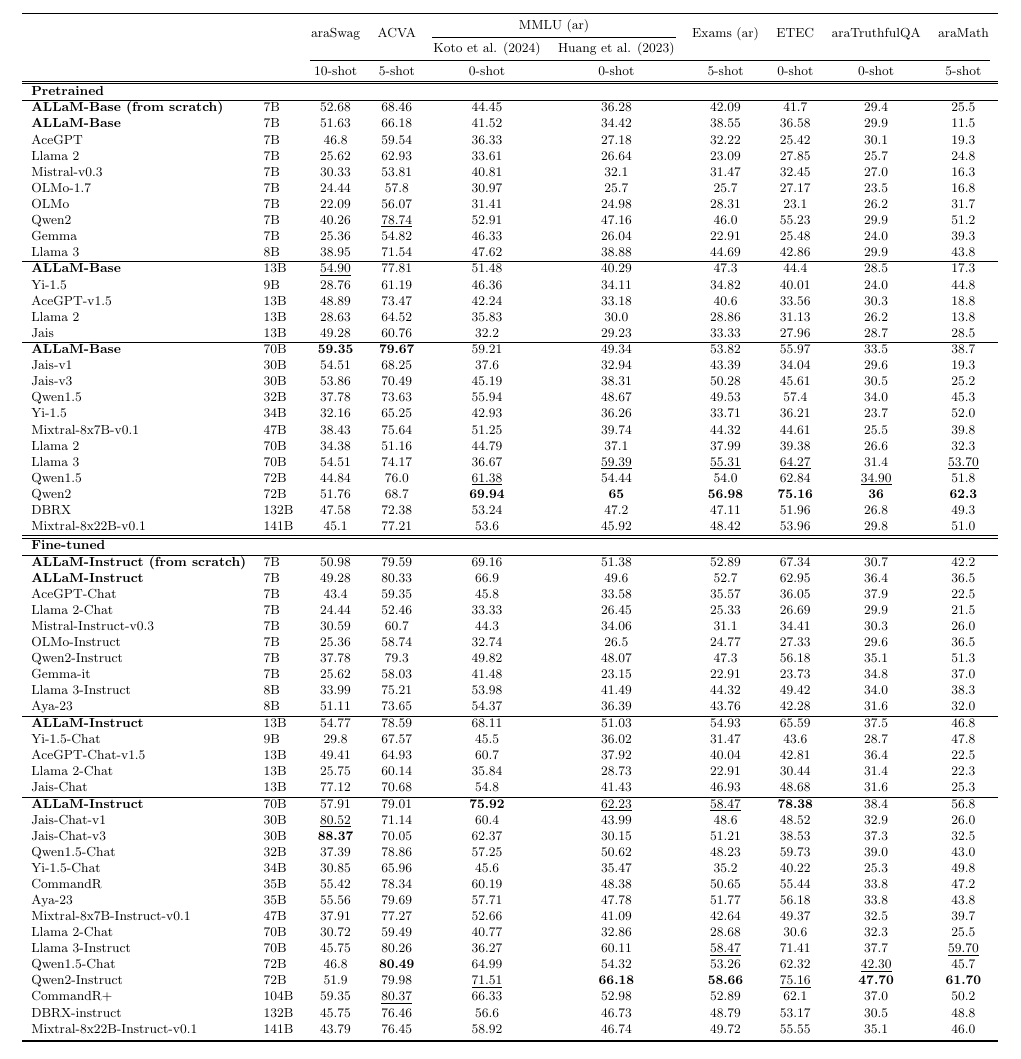}
    \caption{Comprehensive Arabic benchmark results.}
    \label{tab:ar_mega_table_appendix}
\end{table*}

\subsection{Detailed English Evaluation}
\begin{table*}
    \centering
    \includegraphics[width=\textwidth,trim={3cm 2cm 3cm 2cm}]{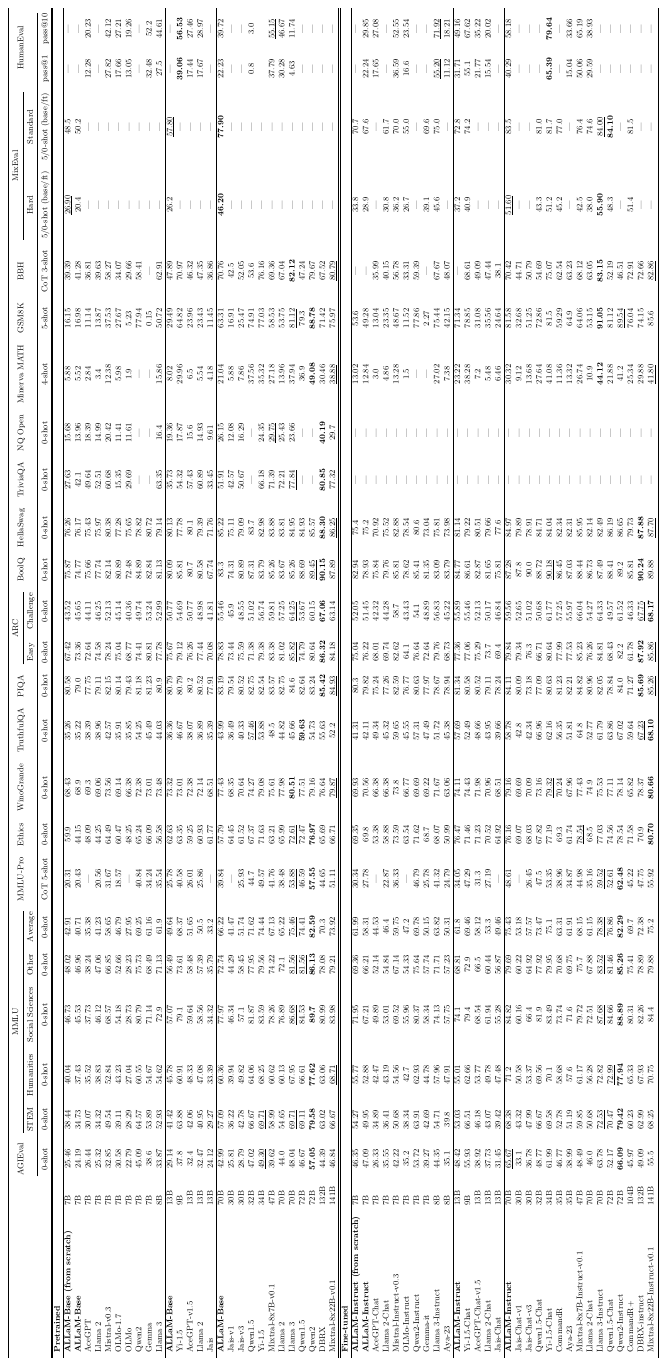}
    \caption{Comprehensive English benchmark results.}
    \label{tab:en_mega_table_appendix}
\end{table*}
\pagebreak
\section{Author Contributions}
In no particular order, 
\begin{itemize}
    \item \textbf{M Saiful Bari:} Research lead for alignment at scale (second language acquisition). Wrote the basic orchestration and optimization of the LLM training framework, data, and training deployment. Trained all the models explained in the paper. Helped build the evaluation infrastructure and orchestrate the entire LLM engine. Implemented the alignment training pipeline and sandboxing environments. Served as a prompt librarian. Supported any project requirements regardless of training, data, or evaluation. Wrote a major part of the paper.
    
    \item \textbf{Yazeed Alnumay:} Implemented pretraining data translation at scale. Worked on vocabulary embeddings augmentation and initialization, as well as the model framework conversion logic. Built the automatic evaluation infrastructure. Acted as a prompt librarian and built the SFT data infrastructure, including SFT and preference data acquisition, curation, processing, cleaning, and lifecycle management. Supported any project requirements regardless of training, data, or evaluation. Contributed to the figures and manuscript writing.
    
    \item \textbf{Haidar Khan}: Guided and reviewed all aspects of the work, with a focus on pretraining and alignment decisions. Contributed directly to optimizing the training infrastructure and trained the 7B model from scratch. Edited and wrote majority of the paper. 

    \item \textbf{Areeb Alowisheq}: Managed the ALLaM training team and wrote the configurations for v0 models. Contributed to all research and logistics decisions for ALLaM. Edited and reviewed the paper, wrote the Arabic LLM related works section. 

    \item \textbf{Nora Al-Twairesh}: Managed the data and evaluation effort of ALLaM. Assisted in building the complete data and evaluation ecosystem for the ALLaM project. Edited and reviewed the paper.
     
    \item \textbf{Abdalghani Abujabal}: Worked on SFT and alignment data creation, quality metrics for data, and model red teaming. Standardized human evaluation for alignment data and led a major part of the data effort. Wrote the data sections of the paper.

    \item \textbf{Jeril Kuriakose}: Worked on MT bench evaluation and maintained all deployment and pretraining data preparation. Supported any project requirements regardless of training, data, or evaluation. Wrote the MT bench evaluation section of the paper. 

    \item \textbf{Ahmed Abdelali}: Worked on the Arabic pretraining data. Led the Arabic OCR effort and book corpus acquisition and creation. Worked on Arabic SFT data.
    
    \item \textbf{Abdulmohsen Al-Thubaity}: Worked on Arabic SFT data creation. Prepared a major part of Arabic SFT data.

    \item \textbf{Zaki Alawami}: Lead the HPC (LLM infrastructure) development and model deployment. 

    \item \textbf{Ali Alammari}: Worked on the acquisition of pretraining Arabic data and the creation of the data processing pipeline.

    \item \textbf{Majed Alrubaian}: Supported the entire data,  training, and evaluation teams. Worked on Arabic Education SFT data. 
    
    \item \textbf{Norah A. Alzahrani}: Implemented all private Arabic evaluations. Lead the human evaluation and prompt acquisition. Maintained the human evaluation pipeline. Wrote the Arabic human evaluation section and prepared the figures. 

    \item \textbf{Nouf M. Alotaibi}:  Trained the v0 version of ALLaM on NeMo. Trained all the tokenizers and prepared training data for the v0 and v3 versions. Prepared the initial SFT figures of the paper. Worked on the red-teaming, SFT quality metrics, and generating visualizations for SFT data.

    \item \textbf{Hisham A. Alyahya}: Trained the v0 version of ALLaM on NeMo, identified bugs in NeMo checkpointing and resuming from checkpoints. Worked on model deployment, measured latency and throughput, and evaluated trade-offs between frameworks like TRT-LLM, vLLM, and HF-TGI. Deployed the preference data collection frontend and wrote the data sampling code for v0 models.

    \item \textbf{Sultan AlRashed}: Worked on Megatron-LM checkpointing and maintained the v0 version of the SFT and RLHF code. Implemented the preference data pipeline in Megatron-LM. Worked on the alignment specialization. Profiled Megatron-Deepspeed.

    \item \textbf{Faisal A. Mirza}: Trained the Arabic-only tokenizer and worked on model creation with an expanded vocabulary.
    
    \item \textbf{Shaykhah Z. Alsubaie}: Maintained evaluation results for the paper, worked on synthetic SFT data creation, specialized model finetuning, few-shot evaluation, and model capability evaluation. Trained router models and worked on the tokenizer. Worked on red-teaming ALLaM models.
    
    \item \textbf{Hassan A. Alahmed}: Worked on pretraining data pipeline for Arabic. 

    \item \textbf{Ghadah Alabduljabbar}: Worked on data pipeline for Arabic, SFT quality metrics. Worked on red-teaming ALLaM models.

    \item \textbf{Raghad Alkhathran}: Worked on data pipeline for Arabic. Worked on red-teaming ALLaM models. 

    \item \textbf{Yousef Almushayqih}: Worked on lm-harness evaluation framework. Worked on red-teaming ALLaM models.

    \item \textbf{Raneem Alnajim}: Worked on training team. Worked on red-teaming ALLaM models.

    \item \textbf{Salman Alsubaihi}: Worked on deployment team. Worked on red-teaming ALLaM models.

    \item \textbf{Maryam Al Mansour}: Worked on the Arabic Data pipeline.
    
\end{itemize}

\paragraph{Note:} In this work we used the \texttt{v2} version of our data and all the model mentioned in the paper are the \texttt{v1} version trained on \texttt{Megatron-LM}. Our \texttt{v0} model series were trained on \texttt{NeMo}.

\end{document}